\documentclass[11pt,a4paper]{article}
\usepackage[hyperref]{acl2020nopatch}
\usepackage{times}
\usepackage{latexsym}

\usepackage{microtype}

\aclfinalcopy 

\usepackage{times}
\usepackage{latexsym}
\usepackage{microtype}
\usepackage{amsfonts, amsmath, amsthm}
\usepackage{makecell}
\usepackage{scalerel}
\usepackage{subfig}

\usepackage{xspace}
\usepackage{url}
\usepackage{xfrac}
\usepackage{booktabs}
\usepackage{tikz}
\usepackage{pgfplots}
\pgfplotsset{compat=1.13}  
\newsavebox{\tempbox}
\usepackage{arydshln}
\usepackage{enumerate}
\usepackage[inline]{enumitem}
\usepackage{float}
\usepackage{adjustbox}

\definecolor{bured}{rgb}{0.8, 0.0, 0.0}

\usepackage{cleveref}
\crefname{section}{\S}{\S\S}
\crefname{table}{Table}{}
\crefname{figure}{Figure}{}
\crefname{algorithm}{Algorithm}{}
\crefname{equation}{eq.}{}
\crefname{appendix}{App.}{}
\crefname{prop}{Proposition}{}
\crefname{thm}{Theorem}{}
\crefformat{section}{\S#2#1#3}  

\newcommand{\sourcelang}{$\mathsf{S}$\xspace}
\newcommand{\targetlang}{$\mathsf{T}$\xspace}
\newcommand{\sourcesent}{S}
\newcommand{\targetsent}{T}
\newcommand{\sourcevocab}{\mathcal{V}_{\mathsf{S}}}
\newcommand{\targetvocab}{\mathcal{V}_{\mathsf{T}}}
\newcommand{\MI}{\mathrm{MI}}
\newcommand{\XMI}{\mathrm{XMI}}

\newcommand{\ent}{\mathrm{H}}
\newcommand{\vs}{\mathbf{s}}
\newcommand{\vt}{\mathbf{t}}

\newcommand{\qlm}{q_{\scriptstyle \mathrm{LM}}}
\newcommand{\qmt}{q_{\scriptstyle \mathrm{MT}}}
\newcommand{\defn}[1]{\textbf{#1}}
\newcommand{\bleu}{\textsc{BLEU}\xspace}
\newcommand{\sacrebleu}{\textsc{SacreBLEU}\xspace}

\everypar{\looseness=-1}

\title{It's Easier to Translate \emph{out of} English than \emph{into} it:\\ Measuring Neural Translation Difficulty by Cross-Mutual Information}

\usepackage{tipa}

\newcommand{\ucambridge}{\text{\normalfont \textipa{D}}}
\newcommand{\ethz}{\text{\normalfont \textipa{Q}}}
\newcommand{\cmu}{\text{\normalfont \textipa{@}}}
\newcommand{\ku}{\text{\normalfont  \textipa{C}}}
\newcommand{\titech}{\text{\normalfont \textipa{N}}}
\newcommand{\jhu}{\text{\normalfont \textipa{H}}}

\author{Emanuele Bugliarello$^\ku$~ Sabrina J. Mielke$^\jhu$~ Antonios Anastasopoulos$^\cmu$ \\
        \textbf{Ryan Cotterell$^{\ucambridge,\ethz}$~ Naoaki Okazaki$^\titech$} \\
        $^\ku$University of Copenhagen~\;~$^\jhu$Johns Hopkins University~\;~$^\cmu$Carnegie Mellon University\\
        $^\ucambridge$University of Cambridge~\;~$^\ethz$ETH Z\"{u}rich~\;~%
        $^\titech$Tokyo Institute of Technology \\
        \texttt{emanuele@di.ku.dk},~\;~ \texttt{sjmielke@jhu.edu},~\;~ \texttt{aanastas@cs.cmu.edu}, \\
        \texttt{rcotterell@inf.ethz.ch},~\;~ \texttt{okazaki@c.titech.ac.jp}
       }

\date{}

\begin{document}
\maketitle

\begin{abstract}
The performance of neural machine translation systems is commonly evaluated in terms of \bleu.
However, due to its reliance on target language properties and generation, the \bleu metric does not allow an assessment of which translation directions are more difficult to model.
In this paper, we propose cross-mutual information ($\XMI$): an asymmetric information-theoretic metric of machine translation difficulty that exploits the probabilistic nature of most neural machine translation models.
$\XMI$ allows us to better evaluate the difficulty of translating text into the target language while controlling for the difficulty of the target-side generation component independent of the translation task.
We then present the first systematic and controlled study of cross-lingual translation difficulties using modern neural translation systems. Code for replicating our experiments is available online at \url{https://github.com/e-bug/nmt-difficulty}.
\end{abstract}

\section{Introduction}
Machine translation (MT) is one of the core research areas in natural language processing. Current state-of-the-art MT systems are based on neural networks \cite{sutskever2014seq2seq,bahdanau2014neural}, which generally surpass phrase-based systems~\cite{koehn2009statistical} in a variety of domains and languages~\cite{bentivogli-etal-2016-neural,toral-sanchez-cartagena-2017-multifaceted,castilho2017neural,bojar-etal-2018-findings,barrault-etal-2019-findings}.
Using phrase-based MT systems, various controlled studies to understand where the translation difficulties lie for different language pairs were conducted \cite{birch-etal-2008-predicting,koehn462}. 
However, comparable studies have yet to be performed for neural machine translation (NMT).
As a result, it is still unclear whether all translation directions are equally easy (or hard) to model for NMT.
This paper hence aims at filling this gap: \textit{Ceteris paribus}, is it easier to translate from English into Finnish or into Hungarian? And how much easier is it? Conversely, is it equally hard to translate Finnish and Hungarian into another language? 
\definecolor{tab1}{rgb}{0.12,0.46,0.70}
\definecolor{tab2}{rgb}{1.0,0.5,0.05}
\begin{figure}
    \centering
    \adjustbox{width=\columnwidth}{
        \begin{tikzpicture}
            \node at (2.77, 5.5) {\textbf{MI}: characterize \emph{language}};
            
            \draw[draw=black,fill=none] (0,0) rectangle (0.6, 4.1) node[pos=.5, rotate=90]{\large $\ent(S)$};
            \draw[draw=none,fill=tab2] (0.9,2.2) rectangle (1.5, 4.1) node[pos=.5, rotate=90, color=white]{$\ent(S \mid T)$};
            \draw[draw=none,fill=tab1] (1.5,0) rectangle (2.1, 2.2) node[pos=.5, rotate=90, color=white]{$\MI(S\,;\,T)$};
            \node at (1, 1.8) {$\Rightarrow$};

            \draw[dashed] (0.9, 2.2) to (4.6, 2.2);
            \node at (2.77, 3.45) {\parbox{6em}{\centering\color{gray} intrinsic source/target language variation}};
            \node at (2.77, 1.05) {\parbox{3em}{\centering\color{gray} shared information}};
            \draw[dashed] (-0.25, 0) to (5.85, 0);
    
            \draw[draw=black,fill=none] (5,0) rectangle (5.6, 4.5) node[pos=.5, rotate=90]{$\ent(T)$};
            \draw[draw=none,fill=tab2] (4.0,2.2) rectangle (4.6, 4.5) node[pos=.5, rotate=90, color=white]{$\ent(T \mid S)$};
            \draw[draw=none,fill=tab1] (3.4,0) rectangle (4.0, 2.2) node[pos=.5, rotate=90, color=white]{$\MI(S\,;\,T)$};
            \node at (4.5, 1.8) {$\Leftarrow$};
            
            
            \begin{scope}[xshift=18em]
            \node at (2.77, 5.5) {\textbf{XMI}: characterize \emph{models}};
            
            \draw[draw=black,fill=none] (0,0) rectangle (0.6, 4.7) node[pos=.5, rotate=90]{\large $\ent_{\qlm}(S)$};
            \draw[draw=none,fill=tab2] (0.9,2.3) rectangle (1.5, 4.9) node[pos=.5, rotate=90, color=white]{$\ent_{\qmt}(S \mid T)$};
            \draw[draw=none,fill=tab1] (1.5,0) rectangle (2.1, 2.5) node[pos=.5, rotate=90, color=white]{$\XMI(T \to S)$};
            \node at (1, 1.8) {$\Rightarrow$};
    
            \node at (2.77, 3.6) {\parbox{6em}{\centering\color{gray} intrinsic source/target modeling difficulty}};
            \node at (2.77, 1.2) {\parbox{3em}{\centering\color{gray} transfer difficulty}};
            \draw[dashed] (-0.25, 0) to (5.85, 0);

            \draw[draw=black,fill=none] (5,0) rectangle (5.6, 5.1) node[pos=.5, rotate=90]{$\ent_{\qlm}(T)$};
            \draw[draw=none,fill=tab2] (4.0,2.5) rectangle (4.6, 4.9) node[pos=.5, rotate=90, color=white]{$\ent_{\qmt}(T \mid S)$};
            \draw[draw=none,fill=tab1] (3.4,0) rectangle (4.0, 2.35) node[pos=.5, rotate=90, color=white]{$\XMI(S \to T)$};
            \node at (4.5, 1.8) {$\Leftarrow$};
            \end{scope}
        \end{tikzpicture}
    }
    \caption{\textbf{Left:} Decomposing the uncertainty of a sentence as mutual information plus language-inherent uncertainty: mutual information ($\MI$) corresponds to just how much easier it becomes to predict $T$ when you are given $S$. $\MI$ is symmetric but the relation between $\ent(S)$ and $\ent(T)$ can be arbitrary. \textbf{Right:} estimating cross-entropies using models $\qmt$ and $\qlm$ invalidates relations between bars, except 
    that $\ent_{q_{\cdot}}(\cdot) \ge \ent(\cdot)$. $\XMI$, our proposed metric, is no longer purely a symmetric measure of language, but now an asymmetric measure that mostly highlights models' shortcomings.} \label{fig:threeblocks}
\end{figure}
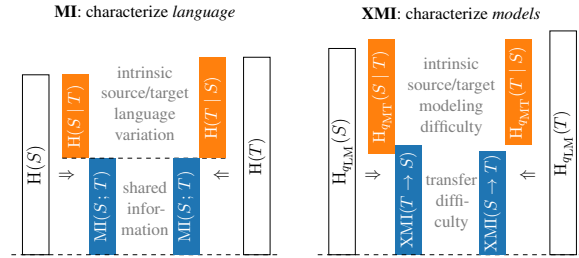

Based on \bleu~\cite{papineni-etal-2002-bleu} scores, previous work~\cite{belinkov-etal-2017-neural} suggests that translating into morphologically rich languages, such as Hungarian or Finnish, is harder than translating into morphologically poor ones, such as English.
However, a major obstacle in the cross-lingual comparison of MT systems is that many automatic evaluation metrics, including \bleu and METEOR~\cite{banerjee-lavie-2005-meteor}, are \emph{not} cross-lingually comparable.
In fact, being a function of $n$-gram overlap between candidate and reference translations, they only allow for a fair comparison of the performance between models when translating into the \emph{same} test set in the \emph{same} target language. 
Indeed, one cannot and should not draw conclusions about the difficulty of translating a source language into different target languages purely based on \bleu (or METEOR) scores.

In response, we propose cross-mutual information ($\XMI$), a new metric towards cross-linguistic comparability in NMT. 
In contrast to \bleu, this information-theoretic quantity no longer explicitly depends on language, model, and tokenization choices. 
It does, however, require that the models under consideration are probabilistic. 
As an initial starting point, we perform a case study with a controlled experiment on $21$ European languages. 
Our analysis showcases $\XMI$'s potential for shedding light on the difficulties of translation as an effect of the properties of the source or target language.
We also perform a correlation analysis in an attempt to further explain our findings. 
Here, in contrast to the general wisdom, we find no significant evidence that translating into a morphologically rich language is harder than translating into a morphologically impoverished one. 
In fact, the only significant correlate
of MT difficulty we find is source-side
type--token ratio.\looseness=-1

\section{Cross-Linguistic Comparability through Likelihoods, not \bleu} \label{sec:not-bleu}
Human evaluation will always be the gold standard of MT evaluation. However, it is both time-consuming and expensive to perform.
To help researchers and practitioners quickly deploy and evaluate new systems, automatic metrics that correlate fairly well with human evaluations have been proposed over the years~\cite{banerjee-lavie-2005-meteor,snover2006study,isozaki-etal-2010-automatic,lo2019yisi}. 
\bleu~\cite{papineni-etal-2002-bleu}, however, has remained the most common metric to report the performance of MT systems.
\bleu is a precision-based metric: a \bleu score is proportional to the geometric average of the number of $n$-grams in the candidate translation that also appear in the reference translation for $1 \leq n \leq 4$.\footnote{\bleu also corrects for reference coverage and includes a length penalty, but we focus on the high-level picture.}

In the context of our study, we take issue with two shortcomings of \bleu scores that prevent a cross-linguistically comparable study. 
First, it is not possible to directly compare \bleu scores across languages because different languages might express the same meaning with a very different number of words.
For instance, agglutinative languages like Turkish often use a single word to express what other languages have periphrastic constructions for. 
To be concrete, the expression ``I will have been programming'' is five words in English, but could easily have been one word in a language with sufficient morphological markings; this unfairly boosts \bleu scores when translating \emph{into} English. 
The problem is further exacerbated by tokenization techniques as finer granularities result in more partial credit and higher $n$ for the $n$-gram matches \citep{post-2018-call}.
In summary, \bleu only allows us to compare models for a \emph{fixed target language and tokenization scheme}, i.e. it only allows us to draw conclusions about the difficulty of translating different source languages into a specific target one (with downstream performance as a proxy for difficulty). 
Thus, \bleu scores cannot provide an answer to which translation direction is easier between \emph{any} two source--target pairs.

In this work, we address this particular shortcoming by considering an information-theoretic evaluation. 
Formally, let $\sourcevocab$ and $\targetvocab$ be source- and target-language vocabularies, respectively.
Let $\sourcesent$ and $\targetsent$ be source- and target-sentence-valued random variables for languages \sourcelang and \targetlang, respectively; then $\sourcesent$ and $\targetsent$ respectively range over $\sourcevocab^*$ and $\targetvocab^*$. 
These random variables $S$ and $T$ are distributed according to some true, unknown probability distribution $p$.
The cross-entropy between the true distribution $p$ and a probabilistic neural translation model $\qmt(\vt \mid \vs)$ is defined as:
\begin{align}\label{eq:xent}
  \ent_{\qmt}(T \mid S) &= \\ 
 & -\sum_{\vt \in \targetvocab^*}\sum_{\vs \in \sourcevocab^*} p(\vt, \vs) \log_2 \qmt(\vt \mid \vs) \nonumber
\end{align}
Since we do not know $p$, we cannot compute \cref{eq:xent}. 
However, given a held-out data set of sentence pairs $\{(\vs^{(i)}, \vt^{(i)})\}_{i=1}^N$ assumed to be drawn from $p$,
we can approximate the true cross-entropy as follows:
\begin{align}\label{eq:empirical-xent}
     \ent_{\qmt}(T &\mid S) \approx \\
     &-\frac{1}{N}\sum_{i=1}^N \log_2 \qmt(\vt^{(i)} \mid \vs^{(i)}) \nonumber
\end{align}
In the limit as $N\rightarrow \infty$, \cref{eq:empirical-xent} converges to \cref{eq:xent}.

We emphasize that this evaluation does not rely on language tokenization provided that the model $\qmt$ does not \cite{Mie2016Can}.
While common in the evaluation of language models, cross-entropy evaluation has been eschewed in machine translation research since (i) not all MT models are probabilistic and (ii) we are often interested in measuring the quality of the candidate translation our model actually produces, e.g. under approximate decoding. 
However, an information-theoretic evaluation is much more suitable for measuring the more abstract
notion of which language pairs are hardest to translate
to and from, which is our purpose here.

\section{Disentangling Translation Difficulty and Monolingual Complexity}\label{sec:nmi}
We contend that simply reporting cross-entropies is not enough.
A second issue in performing a controlled, cross-lingual MT comparison is that the language generation component (without translation) is not equally difficult across languages~\cite{cotterell-etal-2018-languages}.
We claim that the difficulty of \emph{translation} corresponds more closely to the \defn{mutual information} $\MI(S; T)$ between the source and target language, which tells us how much easier it  becomes to predict $T$ when $S$ is given (see \cref{fig:threeblocks}). 
But what is the appropriate analogue of mutual information for cross-entropy?
One such natural generalization is a novel quantity that we term \defn{cross-mutual information}, defined as:
\begin{align}\label{eq:mi}
    \XMI(S \to T) &= \ent_{q_{\mathrm{LM}}}(T) - \ent_{q_{\mathrm{MT}}}(T \mid S)
\end{align}
where $\ent_{q_{\mathrm{LM}}}(T)$ denotes the cross-entropy of the target sentence $\targetsent$ under the model $q_{\mathrm{LM}}$.
As in \cref{sec:not-bleu}, this can, analogously, be approximated by the cross-entropy of a separate target-side language model $\qlm$ over our held-out data set:
\begin{align}\label{eq:ht}
\XMI(S \to T) &\approx \\
&-\frac{1}{N}\sum_{i=1}^N \log_2 \frac{\qlm(\vt^{(i)})}{\qmt(\vt^{(i)}\mid \vs^{(i)})} \nonumber
\end{align}
which, again, becomes exact as $N \rightarrow \infty$. 
In practice, we note that we mix different distributions $\qlm(\vt)$ and $\qmt(\vt \mid \vs)$ and, thus, $\qlm(\vt)$ is \emph{not} necessarily representable as a marginal: there need not be any distribution $\Tilde{q}(\vs)$ such that $\qlm(\vt) = \sum_{\vs \in \sourcevocab^*} \qmt(\vt \mid \vs) \cdot \Tilde{q}(\vs)$. 
While $\qmt$ and $\qlm$ can, in general, be any two models, we exploit the characteristics of NMT models to provide a more meaningful, model-specific estimate of $\XMI$.
NMT architectures typically consist of two components: an encoder that embeds the input text sequence, and a decoder that generates translated output text.
The latter acts as a conditional language model, where the source-language sentence embedded by the encoder drives the target-language generation.
Hence, we use the decoder of $\qmt$ as our $\qlm$ to accurately estimate the difficulty of translation for a given architecture in a controlled way.

In summary, by looking at $\XMI$, we can effectively decouple the language generation component, whose difficulties have been investigated by~\citealt{cotterell-etal-2018-languages} and~\citealt{mielke-etal-2019-kind}, from the translation component. This gives us a measure of how rich and useful the information extracted from the source language is for the target-language generation component.

\section{Experiments}
In order to measure which pairs of languages are harder to translate to and from, we make use of the latest release \texttt{v7} of Europarl~\cite{koehn2005europarl}: a corpus of the proceedings of the European Parliament containing parallel sentences between English (\texttt{en}) and $20$ other European languages: Bulgarian (\texttt{bg}), Czech (\texttt{cs}), Danish (\texttt{da}), German (\texttt{de}), Greek (\texttt{el}), Spanish (\texttt{es}), Estonian (\texttt{et}), Finnish (\texttt{fi}), French (\texttt{fr}), Hungarian (\texttt{hu}), Italian (\texttt{it}), Lithuanian (\texttt{lt}), Latvian (\texttt{lv}), Dutch (\texttt{nl}), Polish (\texttt{pl}), Portuguese (\texttt{pt}), Romanian (\texttt{ro}), Slovak (\texttt{sk}), Slovene (\texttt{sl}) and Swedish (\texttt{sv}). 

\paragraph{Pre-processing steps}
In order to precisely effect a fully controlled experiment, we enforce a \emph{fair} comparison by selecting the set of parallel sentences available across \emph{all} $21$ languages in Europarl. 
This fully controls for the semantic content of the sentences; however, we cannot adequately control for translationese~\cite{stymne-2017-effect,zhang2019effect}.
Our subset of Europarl contains $190{,}733$ sentences for training, $1{,}000$ unique, random sentences for validation and $2{,}000$ unique, random sentences for testing.
For each parallel corpus, we jointly learn byte-pair encodings \cite[BPE;][]{sennrich-etal-2016-neural} for the source and target languages, using $16{,}000$ merge operations. We use the same vocabularies for the language models.\footnote{For English, we arbitrarily chose the English portion of the \texttt{en-bg} vocabulary.}

\begin{table*}
    \centering
    \adjustbox{width=\linewidth}{
        \setlength\tabcolsep{2.5pt}
        \newcommand\dottedcircle{\raisebox{-.1em}{\tikz \draw [line cap=round, line width=0.2ex, dash pattern=on 0pt off 0.5ex] (0,0) circle [radius=0.79ex];}}
        \begin{tabular}{lrrrrrrrrrrrrrrrrrrrrr}
            \toprule
            \phantom{\textbf{en}}\llap{\dottedcircle} $\to$ \textbf{en} & bg & cs & da & de & el & \textbf{es} & et & \colorbox{gray!20}{fi} & fr & hu & it & \textbf{lt} & lv & nl & pl & pt & ro & sk & sl & sv & \textbf{avg}\\
            \midrule
            \bleu & 47.4 & 42.4 & 46.3 & 44.0 & 50.0 & 50.6 & 39.3 & 38.2 & 44.9 & 38.4 & 40.8 & 37.6 & 40.3 & 38.3 & 39.8 & 48.3 & 50.5 & 44.2 & 45.3 & 43.7 & 43.5\\
            $\XMI($\dottedcircle$\,\to\,$\texttt{en}$)$ & 102.3 & 97.0 & 99.7 & 96.5 & 105.3 & 103.8 & 92.8 & 92.1 & 97.0 & 92.5 & 92.1 & 89.2 & 94.2 & 86.5 & 91.9 & 102.5 & 106.1 & 99.8 & 100.1 & 96.9 & 96.9 \\
            $\ent_{\qlm}(\texttt{en})$ & \multicolumn{20}{c}{{\color{gray}\rule[.25em]{25em}{.05em}} 154.2 \color{gray}{\rule[.25em]{25em}{.05em}}} & 154.2 \\
            $\ent_{\qmt}(\texttt{en} \mid \dottedcircle)$ & 51.8 & 57.2 & 54.5 & 57.7 & 48.9 & 50.4 & 61.4 & 62.0 & 57.2 & 61.6 & 62.1 & 65.0 & 60.0 & 67.7 & 62.3 & 51.7 & 48.1 & 54.4 & 54.1 & 57.3 & 57.3 \\
            \toprule
            \textbf{en} $\to$ \phantom{\textbf{en}}\llap{\dottedcircle}\hspace*{1em} & bg & cs & da & de & el & es & et & \colorbox{gray!20}{fi} & fr & hu & it & lt & lv & nl & pl & pt & ro & sk & sl & sv & \textbf{avg}\\
            \midrule
            \bleu &  46.3 &  34.7 &  45.0 &  36.3 &  \colorbox{gray!20}{45.5} &  \colorbox{gray!20}{50.2} &  27.7 &  30.5 &  45.7 &  30.3 &  37.9 &  31.0 &  34.6 &  34.9 &  30.5 &  46.7 &  44.2 &  39.8 &  41.5 &  41.3 & 38.73 \\
            $\XMI($\texttt{en}$\,to\,$\dottedcircle$)$ & 106.2 & 102.8 & 103.3 & 104.0 & \colorbox{gray!20}{111.0} & \colorbox{gray!20}{108.1} & 100.2 & 98.0 & 99.7 & 99.1 & 95.3 & 96.0 & 99.3 & 90.4 & 98.3 & 105.2 & 112.4 & 105.8 & 107.9 & 100.1 & 102.1 \\
            $\ent_{\qlm}(\dottedcircle)$ & 156.5 & 164.0 & 152.7 & 167.6 & 163.7 & 159.3 & 162.5 & 158.6 & 154.9 & 166.6 & 158.6 & 159.2 & 156.4 & 159.7 & 163.4 & 159.3 & 160.5 & 157.7 & 158.2 & 153.1 & 159.6 \\
            $\ent_{\qmt}(\dottedcircle \mid \texttt{en})$ & 50.3 & 61.2 & 49.4 & 63.6 & 52.7 & 51.3 & 62.4 & 60.6 & 55.1 & 67.5 & 63.3 & 63.1 & 57.0 & 69.3 & 65.1 & 54.1 & 48.1 & 51.9 & 50.3 & 53.0 & 57.5 \\
            \bottomrule
        \end{tabular}
    }
    \caption{Test scores, from and into English, Europarl, visualized in \cref{fig:correlations} and \cref{fig:stack}.} \label{tab:from-and-into-en}
\end{table*}

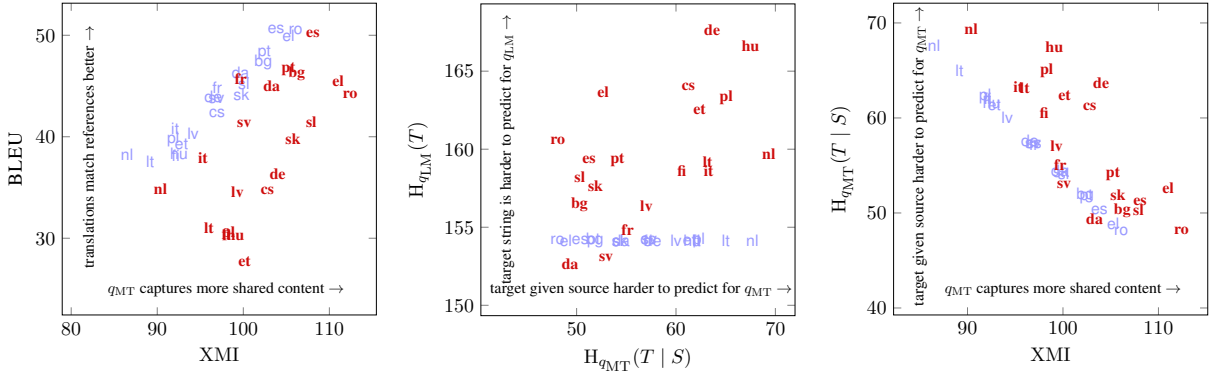
\begin{figure*}[t]
    \center
    \adjustbox{width=\linewidth}{
        \begin{tikzpicture}
            \begin{axis}[
                    xlabel={$\XMI$},
                    ylabel={\bleu},
                    width=20em,
                    height=20em,
                    enlarge x limits=0.1,
                    enlarge y limits=0.1,
                    mark options={scale=0.8}]
                \addplot[mark=text, black, text mark={$\qmt$ captures more shared content $\to$}] coordinates {(98, 25)};
                \addplot[mark=text, black, text mark={\rotatebox{90}{translations match references better $\to$}}] coordinates {(82, 39)};
                \addplot[blue!40, mark=text, text mark={\bf\sf bg}] coordinates {(102.3396, 47.4)};
                \addplot[blue!40, mark=text, text mark={\bf\sf cs}] coordinates {(96.9558, 42.4)};
                \addplot[blue!40, mark=text, text mark={\bf\sf da}] coordinates {(99.6932, 46.3)};
                \addplot[blue!40, mark=text, text mark={\bf\sf de}] coordinates {(96.53, 44.0)};
                \addplot[blue!40, mark=text, text mark={\bf\sf el}] coordinates {(105.2602, 50.0)};
                \addplot[blue!40, mark=text, text mark={\bf\sf es}] coordinates {(103.8174, 50.6)};
                \addplot[blue!40, mark=text, text mark={\bf\sf et}] coordinates {(92.8232, 39.3)};
                \addplot[blue!40, mark=text, text mark={\bf\sf fi}] coordinates {(92.1413, 38.2)};
                \addplot[blue!40, mark=text, text mark={\bf\sf fr}] coordinates {(96.9629, 44.9)};
                \addplot[blue!40, mark=text, text mark={\bf\sf hu}] coordinates {(92.54, 38.4)};
                \addplot[blue!40, mark=text, text mark={\bf\sf it}] coordinates {(92.0796, 40.8)};
                \addplot[blue!40, mark=text, text mark={\bf\sf lt}] coordinates {(89.171, 37.6)};
                \addplot[blue!40, mark=text, text mark={\bf\sf lv}] coordinates {(94.1704, 40.3)};
                \addplot[blue!40, mark=text, text mark={\bf\sf nl}] coordinates {(86.4946, 38.3)};
                \addplot[blue!40, mark=text, text mark={\bf\sf pl}] coordinates {(91.8679, 39.8)};
                \addplot[blue!40, mark=text, text mark={\bf\sf pt}] coordinates {(102.4575, 48.3)};
                \addplot[blue!40, mark=text, text mark={\bf\sf ro}] coordinates {(106.0929, 50.5)};
                \addplot[blue!40, mark=text, text mark={\bf\sf sk}] coordinates {(99.7949, 44.2)};
                \addplot[blue!40, mark=text, text mark={\bf\sf sl}] coordinates {(100.0662, 45.3)};
                \addplot[blue!40, mark=text, text mark={\bf\sf sv}] coordinates {(96.8784, 43.7)};
                \addplot[bured!95, mark=text, text mark={\bf bg}] coordinates {(106.2096, 46.3)};
                \addplot[bured!95, mark=text, text mark={\bf cs}] coordinates {(102.8122, 34.7)};
                \addplot[bured!95, mark=text, text mark={\bf da}] coordinates {(103.3194, 45.0)};
                \addplot[bured!95, mark=text, text mark={\bf de}] coordinates {(104.0, 36.3)};
                \addplot[bured!95, mark=text, text mark={\bf el}] coordinates {(111.0332, 45.5)};
                \addplot[bured!95, mark=text, text mark={\bf es}] coordinates {(108.0881, 50.2)};
                \addplot[bured!95, mark=text, text mark={\bf et}] coordinates {(100.1656, 27.7)};
                \addplot[bured!95, mark=text, text mark={\bf fi}] coordinates {( 98.0227, 30.5)};
                \addplot[bured!95, mark=text, text mark={\bf fr}] coordinates {( 99.7149, 45.7)};
                \addplot[bured!95, mark=text, text mark={\bf hu}] coordinates {( 99.1018, 30.3)};
                \addplot[bured!95, mark=text, text mark={\bf it}] coordinates {( 95.3096, 37.9)};
                \addplot[bured!95, mark=text, text mark={\bf lt}] coordinates {( 96.0001, 31.0)};
                \addplot[bured!95, mark=text, text mark={\bf lv}] coordinates {( 99.3214, 34.6)};
                \addplot[bured!95, mark=text, text mark={\bf nl}] coordinates {( 90.3871, 34.9)};
                \addplot[bured!95, mark=text, text mark={\bf pl}] coordinates {( 98.2996, 30.5)};
                \addplot[bured!95, mark=text, text mark={\bf pt}] coordinates {(105.2407, 46.7)};
                \addplot[bured!95, mark=text, text mark={\bf ro}] coordinates {(112.4211, 44.2)};
                \addplot[bured!95, mark=text, text mark={\bf sk}] coordinates {(105.7748, 39.8)};
                \addplot[bured!95, mark=text, text mark={\bf sl}] coordinates {(107.9064, 41.5)};
                \addplot[bured!95, mark=text, text mark={\bf sv}] coordinates {(100.1222, 41.3)};
            \end{axis}
        \end{tikzpicture}
        \hspace*{.5em}
        \raisebox{-.5em}{
            \begin{tikzpicture}
                \begin{axis}[
                        xlabel={$\ent_{\qmt}(T \mid S)$},
                        ylabel={$\ent_{\qlm}(T)$},
                        width=20em,
                        height=20em,
                        enlarge x limits=0.1,
                        enlarge y limits=0.1,
                        mark options={scale=0.8}]
                    \addplot[mark=text, black, text mark={target given source harder to predict for $\qmt$ $\to$}] coordinates {(56.5, 151)};
                    \addplot[mark=text, black, text mark={\rotatebox{90}{target string is harder to predict for $\qlm$ $\to$}}] coordinates {(43, 160)};
                    \addplot[blue!40, mark=text, text mark={\bf\sf bg}] coordinates{(51.8457, 154.1853)};
                    \addplot[blue!40, mark=text, text mark={\bf\sf cs}] coordinates{(57.2295, 154.1853)};
                    \addplot[blue!40, mark=text, text mark={\bf\sf da}] coordinates{(54.4921, 154.1853)};
                    \addplot[blue!40, mark=text, text mark={\bf\sf de}] coordinates{(57.6553, 154.1853)};
                    \addplot[blue!40, mark=text, text mark={\bf\sf el}] coordinates{(48.9251, 154.1853)};
                    \addplot[blue!40, mark=text, text mark={\bf\sf es}] coordinates{(50.3679, 154.1853)};
                    \addplot[blue!40, mark=text, text mark={\bf\sf et}] coordinates{(61.3621, 154.1853)};
                    \addplot[blue!40, mark=text, text mark={\bf\sf fi}] coordinates{(62.044, 154.1853)};
                    \addplot[blue!40, mark=text, text mark={\bf\sf fr}] coordinates{(57.2224, 154.1853)};
                    \addplot[blue!40, mark=text, text mark={\bf\sf hu}] coordinates{(61.6453, 154.1853)};
                    \addplot[blue!40, mark=text, text mark={\bf\sf it}] coordinates{(62.1057, 154.1853)};
                    \addplot[blue!40, mark=text, text mark={\bf\sf lt}] coordinates{(65.0143, 154.1853)};
                    \addplot[blue!40, mark=text, text mark={\bf\sf lv}] coordinates{(60.0149, 154.1853)};
                    \addplot[blue!40, mark=text, text mark={\bf\sf nl}] coordinates{(67.6907, 154.1853)};
                    \addplot[blue!40, mark=text, text mark={\bf\sf pl}] coordinates{(62.3174, 154.1853)};
                    \addplot[blue!40, mark=text, text mark={\bf\sf pt}] coordinates{(51.7278, 154.1853)};
                    \addplot[blue!40, mark=text, text mark={\bf\sf ro}] coordinates{(48.0924, 154.1853)};
                    \addplot[blue!40, mark=text, text mark={\bf\sf sk}] coordinates{(54.3904, 154.1853)};
                    \addplot[blue!40, mark=text, text mark={\bf\sf sl}] coordinates{(54.1191, 154.1853)};
                    \addplot[blue!40, mark=text, text mark={\bf\sf sv}] coordinates{(57.3069, 154.1853)};
                    \addplot[bured!95, mark=text, text mark={\bf bg}] coordinates{(50.2761, 156.4857)};
                    \addplot[bured!95, mark=text, text mark={\bf cs}] coordinates{(61.2360, 164.0482)};
                    \addplot[bured!95, mark=text, text mark={\bf da}] coordinates{(49.3525, 152.6719)};
                    \addplot[bured!95, mark=text, text mark={\bf de}] coordinates{(63.6453, 167.6453)};
                    \addplot[bured!95, mark=text, text mark={\bf el}] coordinates{(52.6653, 163.6985)};
                    \addplot[bured!95, mark=text, text mark={\bf es}] coordinates{(51.2585, 159.3466)};
                    \addplot[bured!95, mark=text, text mark={\bf et}] coordinates{(62.3698, 162.5354)};
                    \addplot[bured!95, mark=text, text mark={\bf fi}] coordinates{(60.5892, 158.6119)};
                    \addplot[bured!95, mark=text, text mark={\bf fr}] coordinates{(55.1352, 154.8501)};
                    \addplot[bured!95, mark=text, text mark={\bf hu}] coordinates{(67.5041, 166.6059)};
                    \addplot[bured!95, mark=text, text mark={\bf it}] coordinates{(63.2954, 158.605)};
                    \addplot[bured!95, mark=text, text mark={\bf lt}] coordinates{(63.1707, 159.1708)};
                    \addplot[bured!95, mark=text, text mark={\bf lv}] coordinates{(57.0305, 156.3519)};
                    \addplot[bured!95, mark=text, text mark={\bf nl}] coordinates{(69.3437, 159.7308)};
                    \addplot[bured!95, mark=text, text mark={\bf pl}] coordinates{(65.0522, 163.3518)};
                    \addplot[bured!95, mark=text, text mark={\bf pt}] coordinates{(54.104, 159.3447)};
                    \addplot[bured!95, mark=text, text mark={\bf ro}] coordinates{(48.1175, 160.5386)};
                    \addplot[bured!95, mark=text, text mark={\bf sk}] coordinates{(51.9047, 157.6795)};
                    \addplot[bured!95, mark=text, text mark={\bf sl}] coordinates{(50.3312, 158.2376)};
                    \addplot[bured!95, mark=text, text mark={\bf sv}] coordinates{(52.9726, 153.0948)};
                \end{axis}
            \end{tikzpicture}
        }
        \hspace*{.7em}
        \begin{tikzpicture}
            \begin{axis}[
                    xlabel={$\XMI$},
                    ylabel={$\ent_{\qmt}(T \mid S)$},
                    width=20em,
                    height=20em,
                    enlarge x limits=0.1,
                    enlarge y limits=0.1,
                    mark options={scale=0.8}]
                \addplot[mark=text, black, text mark={$\qmt$ captures more shared content $\to$}] coordinates {(100, 42)};
                \addplot[mark=text, black, text mark={\rotatebox{90}{target given source harder to predict for $\qmt$ $\to$}}] coordinates {(85, 56)};
                \addplot[blue!40, mark=text, text mark={\bf\sf bg}] coordinates{(102.3396, 51.8457)};
                \addplot[blue!40, mark=text, text mark={\bf\sf cs}] coordinates{(96.9558, 57.2295)};
                \addplot[blue!40, mark=text, text mark={\bf\sf da}] coordinates{(99.6932, 54.4921)};
                \addplot[blue!40, mark=text, text mark={\bf\sf de}] coordinates{(96.53, 57.6553)};
                \addplot[blue!40, mark=text, text mark={\bf\sf el}] coordinates{(105.2602, 48.9251)};
                \addplot[blue!40, mark=text, text mark={\bf\sf es}] coordinates{(103.8174, 50.3679)};
                \addplot[blue!40, mark=text, text mark={\bf\sf et}] coordinates{(92.8232, 61.3621)};
                \addplot[blue!40, mark=text, text mark={\bf\sf fi}] coordinates{(92.1413, 62.044)};
                \addplot[blue!40, mark=text, text mark={\bf\sf fr}] coordinates{(96.9629, 57.2224)};
                \addplot[blue!40, mark=text, text mark={\bf\sf hu}] coordinates{(92.54, 61.6453)};
                \addplot[blue!40, mark=text, text mark={\bf\sf it}] coordinates{(92.0796, 62.1057)};
                \addplot[blue!40, mark=text, text mark={\bf\sf lt}] coordinates{(89.171, 65.0143)};
                \addplot[blue!40, mark=text, text mark={\bf\sf lv}] coordinates{(94.1704, 60.0149)};
                \addplot[blue!40, mark=text, text mark={\bf\sf nl}] coordinates{(86.4946, 67.6907)};
                \addplot[blue!40, mark=text, text mark={\bf\sf pl}] coordinates{(91.8679, 62.3174)};
                \addplot[blue!40, mark=text, text mark={\bf\sf pt}] coordinates{(102.4575, 51.7278)};
                \addplot[blue!40, mark=text, text mark={\bf\sf ro}] coordinates{(106.0929, 48.0924)};
                \addplot[blue!40, mark=text, text mark={\bf\sf sk}] coordinates{(99.7949, 54.3904)};
                \addplot[blue!40, mark=text, text mark={\bf\sf sl}] coordinates{(100.0662, 54.1191)};
                \addplot[blue!40, mark=text, text mark={\bf\sf sv}] coordinates{(96.8784, 57.3069)};
                \addplot[bured!95, mark=text, text mark={\bf bg}] coordinates{(106.2096, 50.2761)};
                \addplot[bured!95, mark=text, text mark={\bf cs}] coordinates{(102.8122, 61.2360)};
                \addplot[bured!95, mark=text, text mark={\bf da}] coordinates{(103.3194, 49.3525)};
                \addplot[bured!95, mark=text, text mark={\bf de}] coordinates{(104.0, 63.6453)};
                \addplot[bured!95, mark=text, text mark={\bf el}] coordinates{(111.0332, 52.6653)};
                \addplot[bured!95, mark=text, text mark={\bf es}] coordinates{(108.0881, 51.2585)};
                \addplot[bured!95, mark=text, text mark={\bf et}] coordinates{(100.1656, 62.3698)};
                \addplot[bured!95, mark=text, text mark={\bf fi}] coordinates{(98.0227, 60.5892)};
                \addplot[bured!95, mark=text, text mark={\bf fr}] coordinates{(99.7149, 55.1352)};
                \addplot[bured!95, mark=text, text mark={\bf hu}] coordinates{(99.1018, 67.5041)};
                \addplot[bured!95, mark=text, text mark={\bf it}] coordinates{(95.3096, 63.2954)};
                \addplot[bured!95, mark=text, text mark={\bf lt}] coordinates{(96.0001, 63.1707)};
                \addplot[bured!95, mark=text, text mark={\bf lv}] coordinates{(99.3214, 57.0305)};
                \addplot[bured!95, mark=text, text mark={\bf nl}] coordinates{(90.3871, 69.3437)};
                \addplot[bured!95, mark=text, text mark={\bf pl}] coordinates{(98.2996, 65.0522)};
                \addplot[bured!95, mark=text, text mark={\bf pt}] coordinates{(105.2407, 54.104)};
                \addplot[bured!95, mark=text, text mark={\bf ro}] coordinates{(112.4211, 48.1175)};
                \addplot[bured!95, mark=text, text mark={\bf sk}] coordinates{(105.7748, 51.9047)};
                \addplot[bured!95, mark=text, text mark={\bf sl}] coordinates{(107.9064, 50.3312)};
                \addplot[bured!95, mark=text, text mark={\bf sv}] coordinates{(100.1222, 52.9726)};
            \end{axis}
        \end{tikzpicture}
    }
    \caption{Some correlations between metrics in \cref{tab:from-and-into-en}, \textsf{\color{blue!40} into} and \textcolor{bured!95}{\bf from} English. More correlations in \cref{fig:more-correlations}.} \label{fig:correlations}
\end{figure*}

\begin{figure*}[tbh]
	\centering
	\includegraphics[width=\linewidth, trim={0cm 0cm 0cm 0cm}, clip]{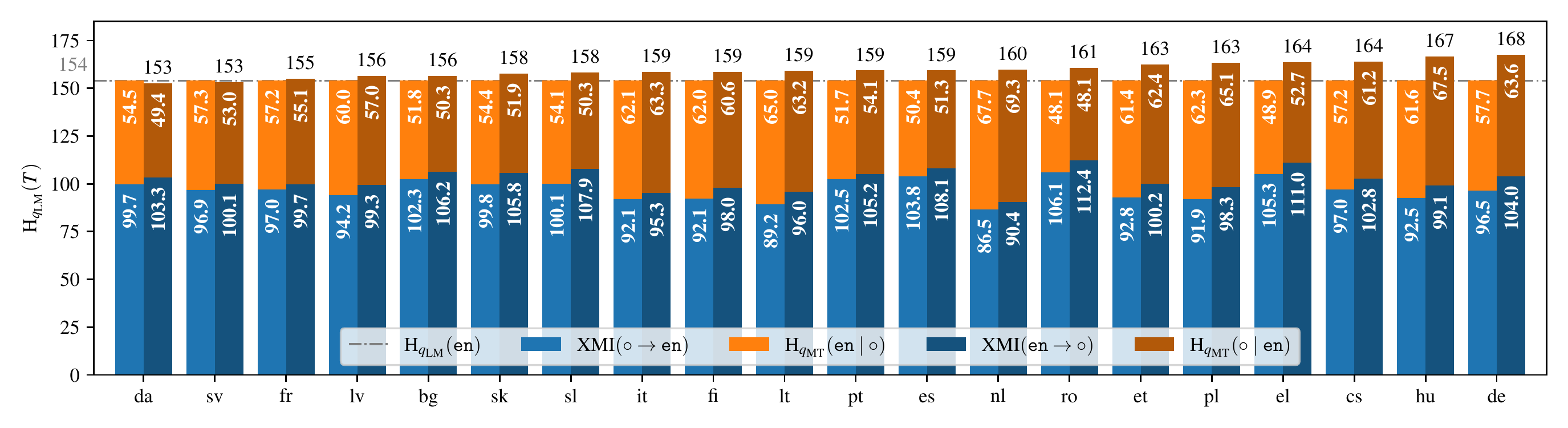}
	\caption{$\ent_{\qlm}(T)$, decomposed into $\XMI(S \to T)$, the information that the system successfully transfers, and $\ent_{\qmt}(T \mid S)$, the uncertainty that remains in the target language, all measured in bits. Note that in $\XMI(S \to T)$ the translation is from the left to the right argument.} \label{fig:stack}
\end{figure*}

\paragraph{Setup}
In our experiments, we train Transformer models~\cite{vaswani2017attention}, which often achieve state-of-the-art performance on MT for various language pairs.
In particular, we rely on the PyTorch~\cite{NEURIPS2019_9015} re-implementation of the Transformer model in the fairseq toolkit~\cite{ott2019fairseq}.
For language modeling, we use the decoder from the same architecture, training it at the sentence level, as opposed to commonly used fixed-length chunks.
We train our systems using label smoothing \cite[LS;][]{szegedy2016rethinking,meister+al.acl20} as it has been shown to prevent models from over-confident predictions, which helps to regularize the models.
We report cross-entropies ($\ent_{\qmt}$, $\ent_{\qlm}$), $\XMI$, and \bleu scores obtained using \sacrebleu~\cite{post-2018-call}.\footnote{Signature: BLEU+c.mixed+\#.1+s.exp+tok.13a+v.1.2.12.}
Finally, in a similar vein to \newcite{cotterell-etal-2018-languages}, we multiply cross-entropy values by the number of sub-word units generated by each model to make our quantities independent of sentence lengths (and divide them by the total number of sentences to match our approximations of the true distributions). 
See~\cref{sec:details} for experimental details.

\section{Results and Analysis}\label{sec:results}
We train $40$ systems, translating each language into and from English.\footnote{Due to resource limitations, we chose these tasks because most of the information available in the web is in English (\url{https://w3techs.com/technologies/overview/content_language}) and effectively translating it into any other language would reduce the digital language divide (\url{http://labs.theguardian.com/digital-language-divide/}). Besides, translating into English gives most people access to any local information.}
The models' performance in terms of \bleu scores, and the cross-mutual information ($\XMI$) and cross-entropy values over the test sets are reported in \cref{tab:from-and-into-en} with significant values marked in \cref{sec:significancetests}.

\paragraph{Translating into English}
When translating into the same target language (in this case, English), \bleu scores are, in fact, comparable, and can be used as a proxy for difficulty. 
We can then conclude, for instance, that Lithuanian (\texttt{lt}) is the hardest language to translate from, while Spanish (\texttt{es}) is the easiest.
In this scenario, given the good correlation of \bleu scores with human evaluations, it is desirable that $\XMI$ correlates well with \bleu. 
This behavior is indeed apparent in the \textsf{\color{blue!40} blue} points in the left part of \cref{fig:correlations}, confirming the efficacy of $\XMI$ in evaluating the difficulty of translation while still being independent of the target language generation component.

\paragraph{Translating from English}
Despite the large gaps between \bleu scores in \cref{tab:from-and-into-en}, one should not be tempted to claim that it is easier to translate into English than from English for these languages as often hinted at in previous work \cite[e.g.,][]{belinkov-etal-2017-neural}.
As we described above, different target languages are not directly comparable, and we actually find that $\XMI$ is slightly higher, on average, when translating from English, indicating that it is actually \emph{easier}, on average, to transfer information correctly in this direction.
For instance, translation from English to Finnish is shown to be easier than from Finnish to English, despite the large gap in \bleu scores. This suggests that the former model is heavily penalized by the target-side language model; this is likely because Finnish has a large number of inflections for nouns and verbs.
Another interesting example is given by Greek (\texttt{el}) and Spanish (\texttt{es}) in \cref{tab:from-and-into-en}, where, again, the two tasks achieve very different \bleu scores but similar $\XMI$.
In light of the correlation with \bleu when translating into English, this shows us that Greek is just harder to language-model, corroborating the findings of~\newcite{mielke-etal-2019-kind}.
Moreover, \cref{fig:correlations} clearly shows that, as expected, $\XMI$ is not as well correlated with \bleu when translating from English, given that \bleu scores are not cross-lingually comparable.

\begin{table}[t]
    \centering
    \footnotesize
    \setlength\tabcolsep{3pt}
    \newcommand\dottedcircle{\raisebox{-.1em}{\tikz \draw [line cap=round, line width=0.2ex, dash pattern=on 0pt off 0.5ex] (0,0) circle [radius=0.79ex];}}
    \begin{tabular}{l|rr}
        \toprule
        \textbf{Metric} & \textbf{Pearson}\hspace*{1.5em} & \textbf{Spearman}\hspace*{1.5em} \\
        \midrule
            word number ratio   & \phantom{-}0.2988 (0.0611)    & \phantom{-}0.3570 (0.0237) \\
            TTR$_{\mathrm{src}}$         & \textbf{-0.5196 (0.0006)}     & \textbf{-0.5136 (0.0007)} \\
            TTR$_{\mathrm{tgt}}$         & \phantom{-}0.1651 (0.3086)    & \phantom{-}0.3355 (0.0343) \\
            $d_{\mathrm{TTR}}$           & -0.4427 (0.0042)     & \textbf{-0.4660 (0.0024)} \\
            word overlap ratio  & \phantom{-}0.1383 (0.3949)    & \phantom{-}0.1731 (0.2853) \\
        \bottomrule
    \end{tabular}
    \caption{Correlation coefficients (and $p$-values) between XMI and data-related features.}
    \label{tab:mini-spearman}
\end{table}

\paragraph{Correlations with linguistic and data features}
Last, we conduct a correlation study between the translation difficulties as measured by $\XMI$ and the linguistic and data-dependent properties of each translation task, following the approaches of \citet{lin-etal-2019-choosing} and \citet{mielke-etal-2019-kind}.
\cref{tab:mini-spearman} lists Pearson's and Spearman's correlation coefficients for data-dependent metrics, where bold values indicate statistically significant results ($p < 0.05$) after Bonferroni correction ($p < 0.0029$).
Interestingly, the only features that significantly correlate with our metric are related to the type-to-token ratio (TTR) for the source language and the distance between source and target TTRs.
This implies that a potential explanation for the differences in translation difficulty lies in lexical variation.
For full correlation results, refer to \cref{sec:pearsons}.

\section{Conclusion}
In this work, we propose a novel information-theoretic approach, $\XMI$, to measure the translation difficulty of probabilistic MT models. Differently from \bleu and other metrics, ours is language- and tokenization-agnostic, enabling the first systematic and controlled study of cross-lingual translation difficulties.
Our results show that $\XMI$ correlates well with \bleu scores when translating into the same language (where they are comparable), and that higher \bleu scores in different languages do not necessarily imply easier translations.
In future work, we plan to extend this analysis across more translation pairs, more diverse languages and multiple domains, as well as investigating the effect of translationese or source-side grammatical errors~\cite{anastasopoulos2019analysis}.

\section*{Acknowledgments}
The authors are thankful to the anonymous reviewers for their valuable feedback. The second-to-last author acknowledges a Facebook Fellowship and discussions
with Tiago Pimentel. 
This project has received funding from the European Union's Horizon 2020 research and innovation programme under the Marie Sk\l{}odowska-Curie grant agreement No 801199, the National Science Foundation under grant 1761548, and by ``Research and Development of Deep Learning Technology for Advanced Multilingual Speech Translation,'' the Commissioned Research of National Institute of Information and Communications Technology (NICT), Japan.

\bibliography{acl2020}
\bibliographystyle{acl_natbib}

\clearpage
\appendix

\section{Experimental Details} \label{sec:details}

\paragraph{Pre-processing steps} 
To precisely determine the effect of the different properties of each language in translation difficulty, we enforce a fair comparison by selecting the same set of parallel sentences across all the languages evaluated in our data set.
The number of parallel sentences available in Europarl varies considerably, ranging from $387K$ sentences for Polish-English to $2.3M$ sentences for Dutch-English.
Therefore, we proceed by taking the set of English sentences that are shared by all the language pairs.
This leaves us with~$197{,}919$ sentences for each language pair, from which we then extract~$1{,}000$ and~$2{,}000$ unique, random sentences for validation and test, respectively.

We follow the same pre-processing steps used by~\newcite{vaswani2017attention} to train the Transformer model on WMT data: Data sets are first tokenized using the Moses toolkit~\cite{koehn-etal-2007-moses} and then filtered by removing sentences longer than $80$ tokens in either source or target language.
Due to this cleaning step that is specific to each training corpus, different sentences are dropped in each data set. 
We then only select the set of sentence pairs that are shared across all languages.
This results in a final number of~190,733 training sentences.
For each parallel corpus, we jointly learn byte-pair encodings (BPE;~\citealp{sennrich-etal-2016-neural}) for source and target languages, using~16,000 merge operations.

\paragraph{Training setup}
In our experiments, we train a Transformer model~\cite{vaswani2017attention}, which achieves state-of-the-art performance on a multitude of language pairs.
In particular, we rely on the PyTorch re-implementation of the Transformer model in the Fairseq toolkit~\cite{ott2019fairseq}.
All experiments are based on the Base Transformer architecture, which we trained for~$20{,}000$ steps and evaluated using the checkpoint corresponding to the lowest validation loss.
We trained our models on a cluster of $4$ machines, each equipped with $4$ Nvidia P100 GPUs, resulting in training times of almost $70$ minutes for each system.
Sentence pairs with similar sequence length were batched together, with each batch containing a total of approximately $32K$ source tokens and $32K$ target tokens.

We used the hyper-parameters specified in latest version ($3$) of Google's Tensor2Tensor~\cite{tensor2tensor} implementation, with the exception of the dropout rate, as we found $0.3$ to be more robust across all the models trained on Europarl.

Models are optimized using Adam~\cite{kingma2014adam} and following the learning schedule specified by~\newcite{vaswani2017attention} with~8,000 warm-up steps.
We employed label smoothing $\epsilon_{ls} = 0.1$~\cite{szegedy2016rethinking} during training and we used beam search with a beam size of $4$ and length penalty $\alpha = 0.6$~\cite{wu2016google}.

For language models, we use a Transformer decoder with the same hyperparameters used in the translation task to effectively measure the contribution given by a translation.
These models were trained, using label smoothing $\epsilon_{ls} = 0.1$, for~10,000 steps on sequences consisting of separate sentences in our corpus.
Analogously to translation models, the checkpoints corresponding to the lowest validation losses were used for evaluation.

\section{Statistical Significance Tests} \label{sec:significancetests}

\begin{table}[t]
	\centering
	\begin{tabular}{l|cc}
		\toprule
        \textbf{Model} & \textbf{Train bootstrap} & \textbf{Test bootstrap} \\
        \midrule
        en-es &  47.6 (0.233) &  50.2 (0.026) \\
        en-et &  25.6 (0.167) &  27.7 (0.026) \\
        lt-en &  34.5 (0.150) &  37.6 (0.027) \\
        ro-en &  47.5 (0.232) &  50.5 (0.027) \\
        \bottomrule
    \end{tabular}
	\caption{Mean test \bleu scores when bootstrapping train and test sets. Numbers in brackets denote standard deviation over $5$ runs (train bootstrap) and $95\%$ confidence interval over $1,000$ samples (test bootstrap).} \label{tab:boot-eu}
\end{table}

\cref{tab:boot-eu} presents the results when applying bootstrap re-sampling~\cite{koehn-2004-statistical} on either training or test sets to the systems achieving the highest and the lowest \bleu scores in the validation set for each direction.
In our experiments, we observe a general trend where the performance of different models varies similarly.
For instance, when we bootstrap test sets, we see that the average \bleu scores are equal to the ones seen in~\cref{tab:from-and-into-en}, and that all the models have similar confidence intervals.\footnote{The same results were observed in all of the $40$ models.}
When bootstrapping the training data, we observe a consistent drop in mean performance of $2-3$ \bleu points across the translation tasks.
The drop in performance is not surprising as the resulting training sets are more redundant, having fewer unique sentences than the original sets, but it is interesting to see that all models are similarly affected. 
The standard deviation over $5$ runs is also similar across all models but slightly larger on the high-performing ones.

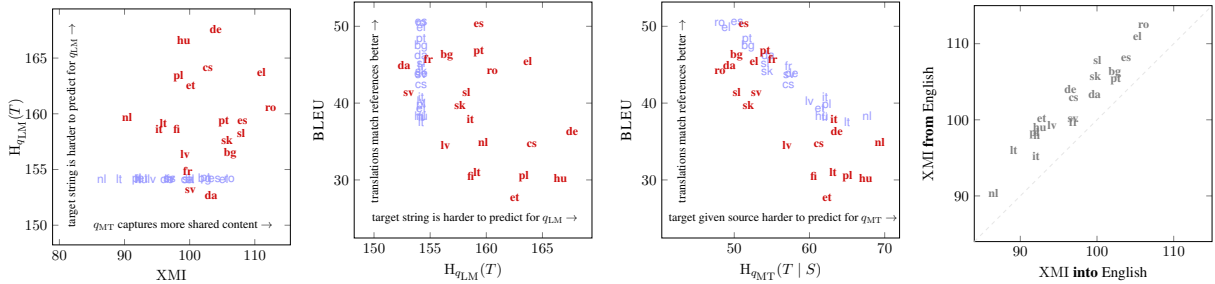
\begin{figure*}[tbh]
    \center
    \adjustbox{width=\linewidth}{
        \begin{tikzpicture}
            \begin{axis}[
                    xlabel={$\XMI$},
                    ylabel={$\ent_{\qlm}(T)$},
                    y label style={at={(axis description cs:-0.09,.5)}},
                    width=20em,
                    height=20em,
                    enlarge x limits=0.1,
                    enlarge y limits=0.1,
                    mark options={scale=0.8}]
                \addplot[mark=text, black, text mark={$\qmt$ captures more shared content $\to$}] coordinates {(99, 150)};
                \addplot[mark=text, black, text mark={\rotatebox{90}{target string is harder to predict for $\qlm$ $\to$}}] coordinates {(82, 159)};
                \addplot[blue!40, mark=text, text mark={\bf\sf bg}] coordinates{(102.3396, 154.1853)};
                \addplot[blue!40, mark=text, text mark={\bf\sf cs}] coordinates{(96.9558, 154.1853)};
                \addplot[blue!40, mark=text, text mark={\bf\sf da}] coordinates{(99.6932, 154.1853)};
                \addplot[blue!40, mark=text, text mark={\bf\sf de}] coordinates{(96.53, 154.1853)};
                \addplot[blue!40, mark=text, text mark={\bf\sf el}] coordinates{(105.2602, 154.1853)};
                \addplot[blue!40, mark=text, text mark={\bf\sf es}] coordinates{(103.8174, 154.1853)};
                \addplot[blue!40, mark=text, text mark={\bf\sf et}] coordinates{(92.8232, 154.1853)};
                \addplot[blue!40, mark=text, text mark={\bf\sf fi}] coordinates{(92.1413, 154.1853)};
                \addplot[blue!40, mark=text, text mark={\bf\sf fr}] coordinates{(96.9629, 154.1853)};
                \addplot[blue!40, mark=text, text mark={\bf\sf hu}] coordinates{(92.54, 154.1853)};
                \addplot[blue!40, mark=text, text mark={\bf\sf it}] coordinates{(92.0796, 154.1853)};
                \addplot[blue!40, mark=text, text mark={\bf\sf lt}] coordinates{(89.171, 154.1853)};
                \addplot[blue!40, mark=text, text mark={\bf\sf lv}] coordinates{(94.1704, 154.1853)};
                \addplot[blue!40, mark=text, text mark={\bf\sf nl}] coordinates{(86.4946, 154.1853)};
                \addplot[blue!40, mark=text, text mark={\bf\sf pl}] coordinates{(91.8679, 154.1853)};
                \addplot[blue!40, mark=text, text mark={\bf\sf pt}] coordinates{(102.4575, 154.1853)};
                \addplot[blue!40, mark=text, text mark={\bf\sf ro}] coordinates{(106.0929, 154.1853)};
                \addplot[blue!40, mark=text, text mark={\bf\sf sk}] coordinates{(99.7949, 154.1853)};
                \addplot[blue!40, mark=text, text mark={\bf\sf sl}] coordinates{(100.0662, 154.1853)};
                \addplot[blue!40, mark=text, text mark={\bf\sf sv}] coordinates{(96.8784, 154.1853)};
                \addplot[bured!95, mark=text, text mark={\bf bg}] coordinates{(106.2096, 156.4857)};
                \addplot[bured!95, mark=text, text mark={\bf cs}] coordinates{(102.8122, 164.0482)};
                \addplot[bured!95, mark=text, text mark={\bf da}] coordinates{(103.3194, 152.6719)};
                \addplot[bured!95, mark=text, text mark={\bf de}] coordinates{(104.0, 167.6453)};
                \addplot[bured!95, mark=text, text mark={\bf el}] coordinates{(111.0332, 163.6985)};
                \addplot[bured!95, mark=text, text mark={\bf es}] coordinates{(108.0881, 159.3466)};
                \addplot[bured!95, mark=text, text mark={\bf et}] coordinates{(100.1656, 162.5354)};
                \addplot[bured!95, mark=text, text mark={\bf fi}] coordinates{(98.0227, 158.6119)};
                \addplot[bured!95, mark=text, text mark={\bf fr}] coordinates{(99.7149, 154.8501)};
                \addplot[bured!95, mark=text, text mark={\bf hu}] coordinates{(99.1018, 166.6059)};
                \addplot[bured!95, mark=text, text mark={\bf it}] coordinates{(95.3096, 158.605)};
                \addplot[bured!95, mark=text, text mark={\bf lt}] coordinates{(96.0001, 159.1708)};
                \addplot[bured!95, mark=text, text mark={\bf lv}] coordinates{(99.3214, 156.3519)};
                \addplot[bured!95, mark=text, text mark={\bf nl}] coordinates{(90.3871, 159.7308)};
                \addplot[bured!95, mark=text, text mark={\bf pl}] coordinates{(98.2996, 163.3518)};
                \addplot[bured!95, mark=text, text mark={\bf pt}] coordinates{(105.2407, 159.3447)};
                \addplot[bured!95, mark=text, text mark={\bf ro}] coordinates{(112.4211, 160.5386)};
                \addplot[bured!95, mark=text, text mark={\bf sk}] coordinates{(105.7748, 157.6795)};
                \addplot[bured!95, mark=text, text mark={\bf sl}] coordinates{(107.9064, 158.2376)};
                \addplot[bured!95, mark=text, text mark={\bf sv}] coordinates{(100.1222, 153.0948)};
            \end{axis}
        \end{tikzpicture}
        \hspace*{.7em}
        \begin{tikzpicture}
            \begin{axis}[
                    xlabel={$\ent_{\qlm}(T)$},
                    ylabel={\bleu},
                    width=20em,
                    height=20em,
                    enlarge x limits=0.1,
                    enlarge y limits=0.1,
                    mark options={scale=0.8}]
                \addplot[mark=text, black, text mark={target string is harder to predict for $\qlm$ $\to$}] coordinates {(159, 25)};
                \addplot[mark=text, black, text mark={\rotatebox{90}{translations match references better $\to$}}] coordinates {(150, 39)};
                \addplot[blue!40, mark=text, text mark={\bf\sf bg}] coordinates{(154.1853, 47.4)};
                \addplot[blue!40, mark=text, text mark={\bf\sf cs}] coordinates{(154.1853, 42.4)};
                \addplot[blue!40, mark=text, text mark={\bf\sf da}] coordinates{(154.1853, 46.3)};
                \addplot[blue!40, mark=text, text mark={\bf\sf de}] coordinates{(154.1853, 44.0)};
                \addplot[blue!40, mark=text, text mark={\bf\sf el}] coordinates{(154.1853, 50.0)};
                \addplot[blue!40, mark=text, text mark={\bf\sf es}] coordinates{(154.1853, 50.6)};
                \addplot[blue!40, mark=text, text mark={\bf\sf et}] coordinates{(154.1853, 39.3)};
                \addplot[blue!40, mark=text, text mark={\bf\sf fi}] coordinates{(154.1853, 38.2)};
                \addplot[blue!40, mark=text, text mark={\bf\sf fr}] coordinates{(154.1853, 44.9)};
                \addplot[blue!40, mark=text, text mark={\bf\sf hu}] coordinates{(154.1853, 38.4)};
                \addplot[blue!40, mark=text, text mark={\bf\sf it}] coordinates{(154.1853, 40.8)};
                \addplot[blue!40, mark=text, text mark={\bf\sf lt}] coordinates{(154.1853, 37.6)};
                \addplot[blue!40, mark=text, text mark={\bf\sf lv}] coordinates{(154.1853, 40.3)};
                \addplot[blue!40, mark=text, text mark={\bf\sf nl}] coordinates{(154.1853, 38.3)};
                \addplot[blue!40, mark=text, text mark={\bf\sf pl}] coordinates{(154.1853, 39.8)};
                \addplot[blue!40, mark=text, text mark={\bf\sf pt}] coordinates{(154.1853, 48.3)};
                \addplot[blue!40, mark=text, text mark={\bf\sf ro}] coordinates{(154.1853, 50.5)};
                \addplot[blue!40, mark=text, text mark={\bf\sf sk}] coordinates{(154.1853, 44.2)};
                \addplot[blue!40, mark=text, text mark={\bf\sf sl}] coordinates{(154.1853, 45.3)};
                \addplot[blue!40, mark=text, text mark={\bf\sf sv}] coordinates{(154.1853, 43.7)};
                \addplot[bured!95, mark=text, text mark={\bf bg}] coordinates{(156.4857, 46.3)};
                \addplot[bured!95, mark=text, text mark={\bf cs}] coordinates{(164.0482, 34.7)};
                \addplot[bured!95, mark=text, text mark={\bf da}] coordinates{(152.6719, 45.0)};
                \addplot[bured!95, mark=text, text mark={\bf de}] coordinates{(167.6453, 36.3)};
                \addplot[bured!95, mark=text, text mark={\bf el}] coordinates{(163.6985, 45.5)};
                \addplot[bured!95, mark=text, text mark={\bf es}] coordinates{(159.3466, 50.2)};
                \addplot[bured!95, mark=text, text mark={\bf et}] coordinates{(162.5354, 27.7)};
                \addplot[bured!95, mark=text, text mark={\bf fi}] coordinates{(158.6119, 30.5)};
                \addplot[bured!95, mark=text, text mark={\bf fr}] coordinates{(154.8501, 45.7)};
                \addplot[bured!95, mark=text, text mark={\bf hu}] coordinates{(166.6059, 30.3)};
                \addplot[bured!95, mark=text, text mark={\bf it}] coordinates{(158.605, 37.9)};
                \addplot[bured!95, mark=text, text mark={\bf lt}] coordinates{(159.1708, 31.0)};
                \addplot[bured!95, mark=text, text mark={\bf lv}] coordinates{(156.3519, 34.6)};
                \addplot[bured!95, mark=text, text mark={\bf nl}] coordinates{(159.7308, 34.9)};
                \addplot[bured!95, mark=text, text mark={\bf pl}] coordinates{(163.3518, 30.5)};
                \addplot[bured!95, mark=text, text mark={\bf pt}] coordinates{(159.3447, 46.7)};
                \addplot[bured!95, mark=text, text mark={\bf ro}] coordinates{(160.5386, 44.2)};
                \addplot[bured!95, mark=text, text mark={\bf sk}] coordinates{(157.6795, 39.8)};
                \addplot[bured!95, mark=text, text mark={\bf sl}] coordinates{(158.2376, 41.5)};
                \addplot[bured!95, mark=text, text mark={\bf sv}] coordinates{(153.0948, 41.3)};
            \end{axis}
        \end{tikzpicture}
        \hspace*{1.1em}
        \begin{tikzpicture}
            \begin{axis}[
                    xlabel={$\ent_{\qmt}(T \mid S)$},
                    ylabel={\bleu},
                    width=20em,
                    height=20em,
                    enlarge x limits=0.1,
                    enlarge y limits=0.1,
                    mark options={scale=0.8}]
                \addplot[mark=text, black, text mark={target given source harder to predict for $\qmt$ $\to$}] coordinates {(56.5, 25)};
                \addplot[mark=text, black, text mark={\rotatebox{90}{translations match references better $\to$}}] coordinates {(43, 39)};
                \addplot[blue!40, mark=text, text mark={\bf\sf bg}] coordinates{(51.8457, 47.4)};
                \addplot[blue!40, mark=text, text mark={\bf\sf cs}] coordinates{(57.2295, 42.4)};
                \addplot[blue!40, mark=text, text mark={\bf\sf da}] coordinates{(54.4921, 46.3)};
                \addplot[blue!40, mark=text, text mark={\bf\sf de}] coordinates{(57.6553, 44.0)};
                \addplot[blue!40, mark=text, text mark={\bf\sf el}] coordinates{(48.9251, 50.0)};
                \addplot[blue!40, mark=text, text mark={\bf\sf es}] coordinates{(50.3679, 50.6)};
                \addplot[blue!40, mark=text, text mark={\bf\sf et}] coordinates{(61.3621, 39.3)};
                \addplot[blue!40, mark=text, text mark={\bf\sf fi}] coordinates{(62.044, 38.2)};
                \addplot[blue!40, mark=text, text mark={\bf\sf fr}] coordinates{(57.2224, 44.9)};
                \addplot[blue!40, mark=text, text mark={\bf\sf hu}] coordinates{(61.6453, 38.4)};
                \addplot[blue!40, mark=text, text mark={\bf\sf it}] coordinates{(62.1057, 40.8)};
                \addplot[blue!40, mark=text, text mark={\bf\sf lt}] coordinates{(65.0143, 37.6)};
                \addplot[blue!40, mark=text, text mark={\bf\sf lv}] coordinates{(60.0149, 40.3)};
                \addplot[blue!40, mark=text, text mark={\bf\sf nl}] coordinates{(67.6907, 38.3)};
                \addplot[blue!40, mark=text, text mark={\bf\sf pl}] coordinates{(62.3174, 39.8)};
                \addplot[blue!40, mark=text, text mark={\bf\sf pt}] coordinates{(51.7278, 48.3)};
                \addplot[blue!40, mark=text, text mark={\bf\sf ro}] coordinates{(48.0924, 50.5)};
                \addplot[blue!40, mark=text, text mark={\bf\sf sk}] coordinates{(54.3904, 44.2)};
                \addplot[blue!40, mark=text, text mark={\bf\sf sl}] coordinates{(54.1191, 45.3)};
                \addplot[blue!40, mark=text, text mark={\bf\sf sv}] coordinates{(57.3069, 43.7)};
                \addplot[bured!95, mark=text, text mark={\bf bg}] coordinates{(50.2761, 46.3)};
                \addplot[bured!95, mark=text, text mark={\bf cs}] coordinates{(61.2360, 34.7)};
                \addplot[bured!95, mark=text, text mark={\bf da}] coordinates{(49.3525, 45.0)};
                \addplot[bured!95, mark=text, text mark={\bf de}] coordinates{(63.6453, 36.3)};
                \addplot[bured!95, mark=text, text mark={\bf el}] coordinates{(52.6653, 45.5)};
                \addplot[bured!95, mark=text, text mark={\bf es}] coordinates{(51.2585, 50.2)};
                \addplot[bured!95, mark=text, text mark={\bf et}] coordinates{(62.3698, 27.7)};
                \addplot[bured!95, mark=text, text mark={\bf fi}] coordinates{(60.5892, 30.5)};
                \addplot[bured!95, mark=text, text mark={\bf fr}] coordinates{(55.1352, 45.7)};
                \addplot[bured!95, mark=text, text mark={\bf hu}] coordinates{(67.5041, 30.3)};
                \addplot[bured!95, mark=text, text mark={\bf it}] coordinates{(63.2954, 37.9)};
                \addplot[bured!95, mark=text, text mark={\bf lt}] coordinates{(63.1707, 31.0)};
                \addplot[bured!95, mark=text, text mark={\bf lv}] coordinates{(57.0305, 34.6)};
                \addplot[bured!95, mark=text, text mark={\bf nl}] coordinates{(69.3437, 34.9)};
                \addplot[bured!95, mark=text, text mark={\bf pl}] coordinates{(65.0522, 30.5)};
                \addplot[bured!95, mark=text, text mark={\bf pt}] coordinates{(54.104, 46.7)};
                \addplot[bured!95, mark=text, text mark={\bf ro}] coordinates{(48.1175, 44.2)};
                \addplot[bured!95, mark=text, text mark={\bf sk}] coordinates{(51.9047, 39.8)};
                \addplot[bured!95, mark=text, text mark={\bf sl}] coordinates{(50.3312, 41.5)};
                \addplot[bured!95, mark=text, text mark={\bf sv}] coordinates{(52.9726, 41.3)};
            \end{axis}
        \end{tikzpicture}
        \hspace*{.7em}
        \begin{tikzpicture}
            \begin{axis}[
                    xlabel={$\XMI$ \textbf{into} English},
                    ylabel={$\XMI$ \textbf{from} English},
                    width=20em,
                    height=20em,
                    enlarge x limits=0,
                    enlarge y limits=0,
                    mark options={scale=0.8}]
                \addplot[mark=none, dashed, black!10] coordinates {(84,84) (115,115)};
                \addplot[gray, mark=text, text mark={\bf bg}] coordinates{(102.3396, 106.2096)};
                \addplot[gray, mark=text, text mark={\bf cs}] coordinates{(96.9558, 102.8122)};
                \addplot[gray, mark=text, text mark={\bf da}] coordinates{(99.6932, 103.3194)};
                \addplot[gray, mark=text, text mark={\bf de}] coordinates{(96.53, 104.0)};
                \addplot[gray, mark=text, text mark={\bf el}] coordinates{(105.2602, 111.0332)};
                \addplot[gray, mark=text, text mark={\bf es}] coordinates{(103.8174, 108.0881)};
                \addplot[gray, mark=text, text mark={\bf et}] coordinates{(92.8232, 100.1656)};
                \addplot[gray, mark=text, text mark={\bf fi}] coordinates{(92.1413, 98.0227)};
                \addplot[gray, mark=text, text mark={\bf fr}] coordinates{(96.9629, 99.7149)};
                \addplot[gray, mark=text, text mark={\bf hu}] coordinates{(92.54, 99.1018)};
                \addplot[gray, mark=text, text mark={\bf it}] coordinates{(92.0796, 95.3096)};
                \addplot[gray, mark=text, text mark={\bf lt}] coordinates{(89.171, 96.0001)};
                \addplot[gray, mark=text, text mark={\bf lv}] coordinates{(94.1704, 99.3214)};
                \addplot[gray, mark=text, text mark={\bf nl}] coordinates{(86.4946, 90.3871)};
                \addplot[gray, mark=text, text mark={\bf pl}] coordinates{(91.8679, 98.2996)};
                \addplot[gray, mark=text, text mark={\bf pt}] coordinates{(102.4575, 105.2407)};
                \addplot[gray, mark=text, text mark={\bf ro}] coordinates{(106.0929, 112.4211)};
                \addplot[gray, mark=text, text mark={\bf sk}] coordinates{(99.7949, 105.7748)};
                \addplot[gray, mark=text, text mark={\bf sl}] coordinates{(100.0662, 107.9064)};
                \addplot[gray, mark=text, text mark={\bf sv}] coordinates{(96.8784, 100.1222)};
            \end{axis}
        \end{tikzpicture}
    }
    \caption{More correlations between metrics in \cref{tab:from-and-into-en}, \textsf{\color{blue!40} into} and \textcolor{bured!95}{\bf from} English.} \label{fig:more-correlations}
\end{figure*}
\begin{table*}[tbh]
    \centering
    \setlength\tabcolsep{3pt}
    \newcommand\dottedcircle{\raisebox{-.1em}{\tikz \draw [line cap=round, line width=0.2ex, dash pattern=on  0pt off  0.5ex] (0,0) circle [radius=0.79ex];}}
    \adjustbox{width=\linewidth}{
        \begin{tabular}{l||cc|c||cc|c}
            \toprule
            \textbf{Metric} & \multicolumn{3}{c}{\textbf{Pearson}} & \multicolumn{3}{c}{\textbf{Spearman}} \\
            & \dottedcircle $\to$ \textbf{en} & \textbf{en} $\to$ \dottedcircle & \textbf{both} & \dottedcircle $\to$ \textbf{en} & \textbf{en} $\to$ \dottedcircle & \textbf{both} \\
            \midrule
            MCC$_{src}$         & \textcolor{gray}{-0.2579 (0.2723)} & \textcolor{gray}{--} & -0.4302 (0.0056)  & \textcolor{gray}{-0.2135 (0.3660)} & \textcolor{gray}{--} & -0.4444 (0.0041) \\
            MCC$_{tgt}$         & \textcolor{gray}{--} & \textcolor{gray}{-0.1260 (0.5965)} & \textcolor{gray}{\phantom{-}0.2619 (0.1025)}   & \textcolor{gray}{--} & \textcolor{gray}{-0.1263 (0.5957)} & \phantom{-}0.3778 (0.0162) \\
            ADL$_{src}$         & \textcolor{gray}{-0.2972 (0.2032)} & \textcolor{gray}{--} & \textcolor{gray}{-0.1166 (0.4737)}  & \textcolor{gray}{-0.2887 (0.2170)} & \textcolor{gray}{--} & \textcolor{gray}{\phantom{-}0.0166 (0.9188)} \\
            ADL$_{tgt}$         & \textcolor{gray}{--} & \textcolor{gray}{-0.2254 (0.3393)} & \textcolor{gray}{-0.2110 (0.1912)}  & \textcolor{gray}{--} & \textcolor{gray}{-0.1820 (0.4426)} & -0.3798 (0.0156) \\
            HPE-mean$_{src}$    & \textcolor{gray}{\phantom{-}0.2012 (0.3950)} & \textcolor{gray}{--} & \phantom{-}0.4567 (0.0031)    & \textcolor{gray}{\phantom{-}0.2000 (0.3979)} & \textcolor{gray}{--}  & \phantom{-}0.4508 (0.0035) \\
            HPE-mean$_{tgt}$    & \textcolor{gray}{--} & \textcolor{gray}{\phantom{-}0.0142 (0.9525)} & -0.4115 (0.0083)   & -- & \textcolor{gray}{\phantom{-}0.0120 (0.9599)}  & -0.4103 (0.0085) \\
            \midrule
            genetic             & \textcolor{gray}{\phantom{-}0.0433 (0.8563)}  & \textcolor{gray}{\phantom{-}0.0777 (0.7446)} & \textcolor{gray}{\phantom{-}0.0544 (0.7387)}   & \textcolor{gray}{-0.1526 (0.5207)}  & \textcolor{gray}{-0.1741 (0.4630)} & \textcolor{gray}{-0.1360 (0.4028)} \\
            syntactic           & \textcolor{gray}{-0.3643 (0.1143)}  & \textcolor{gray}{-0.2056 (0.3845)} & \textcolor{gray}{-0.2556 (0.1114)}                       & \textcolor{gray}{-0.3560 (0.1234)}  & \textcolor{gray}{-0.2695 (0.2506)} & \textcolor{gray}{-0.2688 (0.0935)} \\
            featural            & \textcolor{gray}{-0.0561 (0.8142)}  & \textcolor{gray}{-0.0577 (0.8090)} & \textcolor{gray}{-0.0511 (0.7540)}                       & \textcolor{gray}{\phantom{-}0.0121 (0.9597)}  & \textcolor{gray}{-0.0093 (0.9690)} & \textcolor{gray}{-0.0109 (0.9467)} \\
            phonological        & \textcolor{gray}{-0.1442 (0.5441)}  & \textcolor{gray}{-0.2222 (0.3465)} & \textcolor{gray}{-0.1647 (0.3097)}                       & \textcolor{gray}{-0.0435 (0.8556)}  & \textcolor{gray}{-0.0948 (0.6909)} & \textcolor{gray}{-0.0906 (0.5782)} \\
            inventory           & \textcolor{gray}{\phantom{-}0.1125 (0.6369)}  & \textcolor{gray}{\phantom{-}0.1048 (0.6601)} & \textcolor{gray}{\phantom{-}0.0976 (0.5492)}     & \textcolor{gray}{\phantom{-}0.1231 (0.6052)}    & \textcolor{gray}{\phantom{-}0.1472 (0.5356)} & \textcolor{gray}{\phantom{-}0.1128 (0.4884)} \\
            geographic          & \textcolor{gray}{\phantom{-}0.1983 (0.4019)}  & \textcolor{gray}{\phantom{-}0.3388 (0.1440)} & \textcolor{gray}{\phantom{-}0.2416 (0.1332)}     & \textcolor{gray}{\phantom{-}0.1336 (0.5745)}    & \textcolor{gray}{\phantom{-}0.2550 (0.2779)} & \textcolor{gray}{\phantom{-}0.2062 (0.2017)} \\
            \midrule
            word number ratio   & \phantom{-}0.4559 (0.0434)  & \textcolor{gray}{-0.2953 (0.2063)} & \textcolor{gray}{\phantom{-}0.2988 (0.0611)}           & \phantom{-}0.4602 (0.0412) &  \textcolor{gray}{-0.3278 (0.1582)} & \phantom{-}0.3570 (0.0237) \\
            TTR$_{src}$             & -0.4746 (0.0345)  & \textcolor{gray}{--}     & \textbf{-0.5196 (0.0006)}                    & -0.4857 (0.0299) & \textcolor{gray}{--}               & \textbf{-0.5136 (0.0007)} \\
            TTR$_{tgt}$             & \textcolor{gray}{--}                & \textcolor{gray}{-0.2931 (0.2099)} & \textcolor{gray}{\phantom{-}0.1651 (0.3086)}           & \textcolor{gray}{--}                   & \textcolor{gray}{-0.3128 (0.1794)} & \phantom{-}0.3355 (0.0343) \\
            $d_{\mathrm{TTR}}$               & \textcolor{gray}{-0.4434 (0.0502)}  & \textcolor{gray}{-0.2404 (0.3072)} & -0.4427 (0.0042)         & -0.4857 (0.0299) & \textcolor{gray}{-0.3128 (0.1794)} & \textbf{-0.4660 (0.0024)} \\
            word overlap ratio       & \textcolor{gray}{\phantom{-}0.2563 (0.2754)}  & \textcolor{gray}{\phantom{-}0.0526 (0.8258)} & \textcolor{gray}{\phantom{-}0.1383 (0.3949)}     & \textcolor{gray}{\phantom{-}0.1474 (0.5352)}    & \textcolor{gray}{\phantom{-}0.1474 (0.5352)} & \textcolor{gray}{\phantom{-}0.1731 (0.2853)} \\
            \bottomrule
        \end{tabular}
    }
    \caption{All Pearson's and Spearman's correlation coefficients and corresponding $p$-values (in brackets) between XMI and various metrics. Values in black are statistically significant at $p<0.05$, and bold values are also statistically significant after Bonferroni correction ($p<0.0029$).}
    \label{tab:all-pearson-spearman}
\end{table*}

\section{More Correlations between Metrics}\label{sec:more-correlations}

\Cref{fig:more-correlations} shows more correlations between the metrics we reported in our experiments (see \cref{tab:from-and-into-en}).

\section{Correlation Analysis}\label{sec:pearsons}

\Cref{tab:all-pearson-spearman} shows Pearson's and Spearman's correlations between $\XMI$ and all investigated predictors, including per-direction results.
Following \citet{lin-etal-2019-choosing} and \citet{mielke-etal-2019-kind}, we evaluated:
\begin{itemize}[noitemsep,topsep=1pt]
    \item MCC: Morphological counting complexity~\cite{sagot-2013-mcc}, using the values for Europarl reported by~\citet{cotterell-etal-2018-languages}.
    \item ADL: Average dependency length~\cite{futrell2015large}, using the values reported for Europarl by~\citet{mielke-etal-2019-kind}.
    \item HPE-mean: mean over all Europarl tokens of Head-POS Entropy~\cite{dehouck-denis-2018-framework}, as reported by~\citet{mielke-etal-2019-kind}.
    \item Six different linguistic distances (genetic, syntactic, featural, phonological, inventory, geographic) from the URIEL Typological Database~\cite{littell-etal-2017-uriel}. We refer the reader to~\citet{lin-etal-2019-choosing} for more details.
    \item Word number ratio: number of source tokens over number of target tokens used for training.
    \item TTR$_{src}$ and TTR$_{tgt}$: type-to-token ratio evaluated on the source and target language training data, respectively, to measure lexical diversity.
    \item $d_{\mathrm{TTR}}$: distance between the TTRs of the source and target language corpora, as a rough indication of their morphological similarity:
    \begin{align*}
        d_{\mathrm{TTR}} = \left(1 - \frac{\mathrm{TTR}_{src}}{\mathrm{TTR}_{tgt}}\right)^2.
    \end{align*}
    \item Word overlap ratio: we measure the similarity between the vocabularies of source and target languages as the ratio between the number of shared types and the size of their union.
\end{itemize}

\end{document}